\documentclass{article}
\usepackage{amssymb}
\usepackage{amssymb}
\usepackage{amsfonts}
\usepackage{mathtools}
\usepackage{bm}
\usepackage{dsfont}

\usepackage[linesnumbered,ruled,vlined]{algorithm2e}

\SetCommentSty{mycommfont}
\SetAlCapFnt{\small}
\SetAlgorithmName{Algo}{Algo}{}
\usepackage[utf8]{inputenc}
\usepackage{booktabs}
\usepackage{multirow}
\usepackage{arydshln}
\usepackage{threeparttable}
\usepackage{float}

\usepackage{graphicx}
\usepackage{wrapfig}

\usepackage[numbers]{natbib}
\usepackage{hyperref}
\usepackage{cleveref}

\usepackage{times}
\usepackage{multicol}
\usepackage{xspace}
\usepackage{xurl}
\usepackage{lipsum}
\usepackage{siunitx} 
\usepackage{balance} 
\usepackage{stfloats}
\usepackage{capt-of}  

\usepackage{stfloats}

\usepackage[dvipsnames,table,xcdraw]{xcolor}
\definecolor{lightgray}{gray}{0.9}
\definecolor{mydarkblue}{rgb}{0,0.08,0.45}
\definecolor{mydarkgreen}{RGB}{0, 139, 69}
\definecolor{mygreen2}{RGB}{0 205 0}
\definecolor{mybrown}{RGB}{139 69 19}
\definecolor{boxblue}{RGB}{79,173,234}
\definecolor{boxgreen}{RGB}{159,206,99}
\definecolor{tablepeach}{RGB}{255, 240, 235}
\definecolor{tablepurple}{RGB}{248,235,252}
\definecolor{tableblue}{RGB}{235,241,255}
\definecolor{lowerbody}{RGB}{76,123,49}
\definecolor{upperbody}{RGB}{47,110,186}

\usepackage{pifont}
\usepackage{bbm}
\usepackage{etoc}    
\usepackage{caption}

\newcommand{\ie}{\emph{i.e. }}
\newcommand{\eg}{\emph{e.g., }}

\newcommand{\methodnosapce}{JAEGER}
\newcommand{\method}{JAEGER\xspace}
\newcommand{\methodours}{JAEGER (Ours)\xspace}
\newcommand{\methodgradient}{\gradientRGB{\methodnosapce}{0,150,255}{0,255,50}}
\usepackage{wrapfig}

\usepackage{gradient-text}

\usepackage{hyperref}
\usepackage{float}
\usepackage{etoolbox}

\usepackage[preprint]{corl_2025} 

\usepackage{enumitem}

\author{
    Ziluo Ding\textsuperscript{\rm 1}$^{*}$\hspace{5pt}
    Haobin Jiang\textsuperscript{\rm 2}$^{*}$\hspace{5pt}
    Yuxuan Wang\textsuperscript{\rm 2}$^{*}$\hspace{5pt}
    Zhenguo Sun\textsuperscript{\rm 1}\hspace{5pt}
    Yu Zhang\textsuperscript{\rm 1}\hspace{5pt}
    Xiaojie Niu\textsuperscript{\rm 1} \\
    \textbf{Ming Yang}\textsuperscript{\rm 2}\hspace{5pt}
    \textbf{Weishuai Zeng}\textsuperscript{\rm 2}\hspace{5pt}
    \textbf{Xinrun Xu}\textsuperscript{\rm 1}\hspace{5pt}
    \textbf{Zongqing Lu}\textsuperscript{\rm 2}$^\dag$ \\
    \\
    \textsuperscript{\rm 1} BAAI \hspace{5pt}
    \textsuperscript{\rm 2} Peking University
}

\newcommand{\insertfig}{
\includegraphics[width=.9\textwidth] {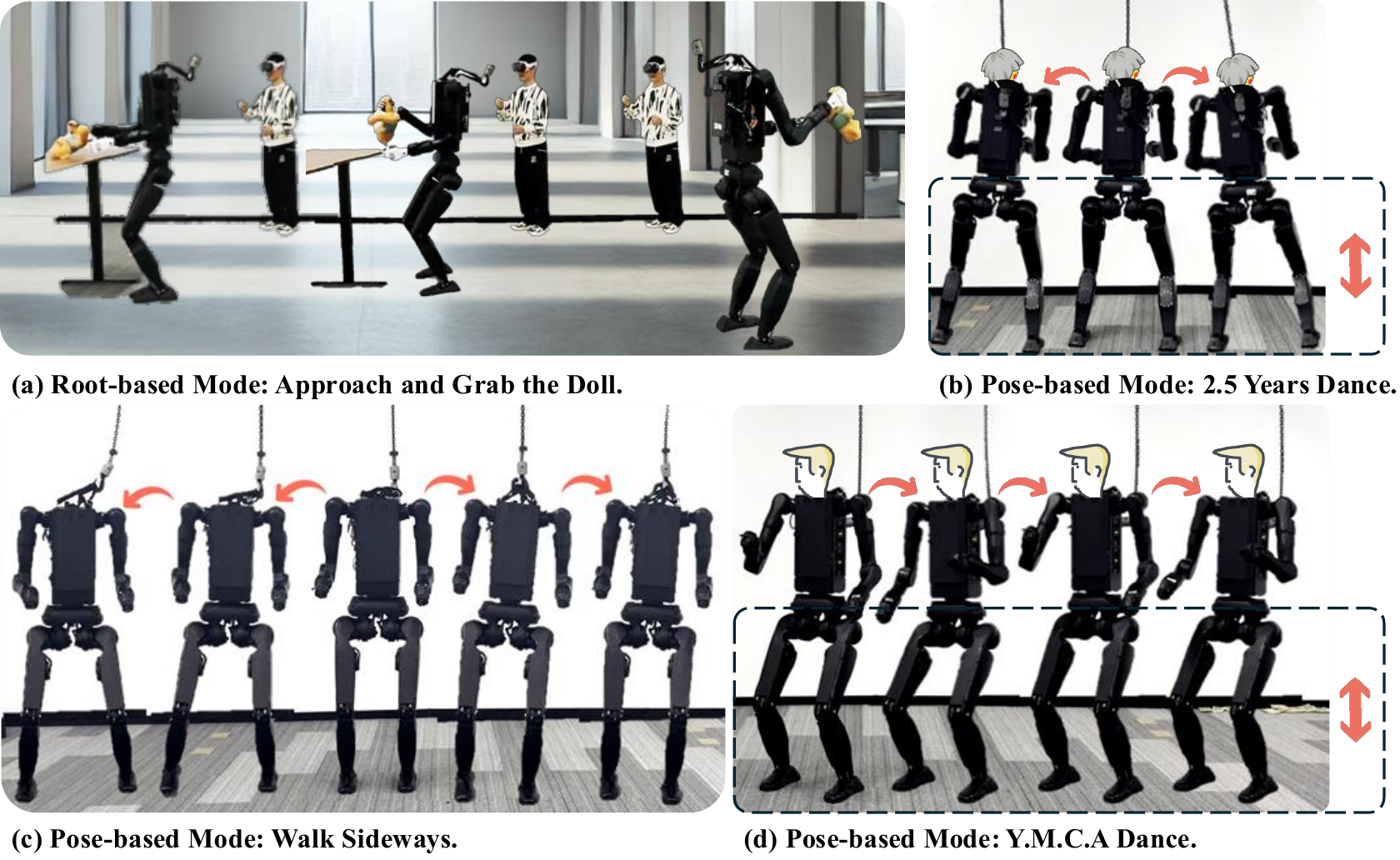} \captionof{figure}{ 
Some real-world demonstrations of \method deployed on the H1-2. For the root-based mode, we use a VR and a joystick to control the upper and lower body of a humanoid for grasping objects while in motion \textbf{(a)}. In the pose-based mode, the robot can perform various dance movements \textbf{(b, d)}. It can also execute large-amplitude lower-body movements, such as walking forward and sideways \textbf{(c)}. }\label{fig:001} \vspace{-0.1cm}}

\newcommand{\inserthref}{
\vspace{-1.3cm}
\begin{center}
\href{https://beingbeyond.github.io/Jaeger}{\texttt{\textcolor{green!70!black}{https://beingbeyond.github.io/Jaeger}}}
\end{center}
}

\makeatletter
\apptocmd{\@maketitle}{\inserthref}{}{}
\apptocmd{\@maketitle}{\centering\insertfig}{}{}
\makeatother

\begin{document}

\title{\methodgradient \\ Dual-Level Humanoid Whole-Body Controller}
\maketitle

\begingroup
\renewcommand\thefootnote{}\footnotetext{$^*$These authors contributed equally to this work.}
\renewcommand\thefootnote{}\footnotetext{$^\dag$Correspondence to Zongqing Lu $<$zongqing.lu@pku.edu.cn$>$.}
\endgroup


\begin{abstract}
This paper presents \method, a dual-level whole-body controller for humanoid robots that addresses the challenges of training a more robust and versatile policy. Unlike traditional single-controller approaches, \method separates the control of the upper and lower bodies into two independent controllers, so that they can better focus on their distinct tasks. This separation alleviates the dimensionality curse and improves fault tolerance. \method supports both root velocity tracking (coarse-grained control) and local joint angle tracking (fine-grained control), enabling versatile and stable movements. To train the controller, we utilize a human motion dataset (AMASS), retargeting human poses to humanoid poses through an efficient retargeting network, and employ a curriculum learning approach. This method performs supervised learning for initialization, followed by reinforcement learning for further exploration. We conduct our experiments on two humanoid platforms and demonstrate the superiority of our approach against state-of-the-art methods in both simulation and real environments. 
\end{abstract}



\section{Introduction}

Due to hardware constraints and the inherent complexity of the robotic action space, achieving effective whole-body control (WBC) for adult-sized humanoid robots, such as the Unitree H1-2, remains a significant challenge. Recent studies on WBC have demonstrated promising advancements, enabling humanoid robots to perform versatile motions by learning from extensive human data \citep{he2024learning,fu2024humanplus,he2024omnih2o,ji2024exbody2,he2024hover}. Based on different task settings, WBC methodologies can be broadly categorized into three types: root velocity tracking \citep{cheng2024express}, kinematic position tracking \citep{he2024learning,he2024omnih2o}, and local joint angle tracking \citep{cheng2024express,fu2024humanplus,ji2024exbody2}. Root velocity tracking emphasizes \emph{coarse-grained control}, where the robot tracks a given velocity \emph{without} relying on a specific reference pose. In contrast, kinematic position and local joint angle tracking focus on accurately reproducing a given trajectory of reference poses, which can be regarded as \emph{fine-grained control} for humanoids. Note that both the kinematic position and the local joint angle can be used to describe the pose of the humanoid at a given moment.

Intuitively, a more versatile WBC controller should integrate both coarse-grained and fine-grained commands, regardless of whether they are represented through positions or angles. However, \textit{the challenge of effectively combining these two approaches remains unresolved.} Current approaches, such as OmniH2O~\cite{he2024omnih2o} and HumanPlus~\cite{fu2024humanplus}, focus on learning locomotion by following reference pose trajectories and deriving velocity tracking capabilities from these trajectories. However, in practice, since the inevitable differences between embodiments of humanoid robots and humans, locomotion by fully imitating human pose might hinder humanoids from maintaining balance and prevent them from learning a robust policy with the capability of velocity tracking. Therefore, we propose to first separately train a root velocity tracking policy and a local joint angle tracking policy, and then distill the two policies into a unified WBC policy.

Given the complexity of the WBC problem and the heterogeneous functionalities of the upper and lower bodies, we reformulate the WBC problem as \textit{a multi-agent system} \citep{lowe2017multi,rashid2020monotonic,yu2022surprising}. During training, we observed that interactions between the upper and lower bodies often hinder convergence; for instance, the upper body’s tracking behavior becomes overly conservative to accommodate the lower body’s stability. Addressing this interference typically requires intricate parameter tuning and curriculum learning. Inspired by the multi-agent system, we propose decoupling the upper and lower bodies into two independent controllers, enabling each to focus on its specific tasks while alleviating the curse of dimensionality in action space \citep{gronauer2022multi}. This design not only simplifies the training process but also enhances the system’s fault tolerance, allowing the lower body to continue functioning effectively even if the upper body encounters failure.

In this paper, we present \method, a dual-level humanoid whole-body controller that supports both root velocity tracking, \ie coarse-grained control, and local joint angle tracking, \ie fine-grained control, as illustrated in Fig. \ref{fig:method_framework}. Given the upper and lower bodies serve different functions, we use two controllers to manage them separately for a more robust WBC policy. To achieve effective coordination, the two controllers share both observations and rewards. For the training procedure, we follow recent success in leveraging a human motion dataset, AMASS~\cite{AMASS:ICCV:2019}. We first retarget human poses to humanoid poses via our efficient retargeting network and then train the two controllers to follow these humanoid poses. Given the inherent challenges of WBC training, we adopt a curriculum learning approach. Initially, the lower-body controller is trained independently. Next, we implement a supervised initialization for the upper-body controller, followed by the whole-body control training using reinforcement learning (RL).

To summarize, our contributions are as follows: 
1) We propose an MLP-based retargeting approach that outperforms optimization-based inverse kinematics (IK) methods by producing more accurate and smoother joint angles for humanoid control, while operating at a high execution frequency of up to 1 kHz. 2) We introduce a novel dual-level WBC controller, which reduces mutual interference between the upper and lower bodies while enabling each controller to focus more on its respective task. 3) To train the dual-level controller, we propose a structured curriculum learning strategy. A tailored supervised initialization serves as the first stage before RL, which facilitates the following RL convergence to a more optimal solution.

\section{Related work}

\subsection{Humanoid Whole-Body Control}
Whole-body control for humanoid robots remains a challenging problem due to hardware limitations and the high complexity of the action space. As pioneers in this field, the H2O series \cite{he2024omnih2o,he2024learning} fully rely on global keypoint position tracking. However, the global keypoint positions may gradually drift over time, leading to cumulative errors that ultimately impede learning. HumanPlus \cite{fu2024humanplus}, on the other hand, fully relies on joint angle tracking to learn motion and autonomous skills from human data. Exbody series combines the strengths of both approaches. In more detail, ExBody \cite{cheng2024express} expresses the diverse upper-body motion by tracking the joint angles and keypoint positions, while tracking the root velocity for the lower body. ExBody2 \cite{ji2024exbody2} achieves superior performance by only learning from a curated subset of the AMASS dataset. It shows the poses that are lower-body feasible are crucial for training stable behaviors. In addition, it claims that the local keypoint position tracking can lead to a more robust strategy. HOVER \cite{he2024hover}, as a unified framework, integrates all the interfaces, \eg root velocity, joint angles, and keypoint positions, into a single network through a distillation process. Our work is concurrent with HOVER and shares some similarities, particularly in the integration of root velocity and joint angle control. However, we train this more challenging controller from a different perspective, namely by decoupling the upper and lower bodies. 

\subsection{Humanoid Locomotion}
Training a robust locomotion model is the core task for the humanoid. With the rise of learning algorithms, reinforcement learning (RL) and imitation learning (IL) have proven to be more robust than traditional model-based control methods~\cite{zhuang2024humanoid, long2024learning, agarwal2023legged}. By designing effective reward functions, RL has shown promise in enabling humanoids to learn complex locomotion skills. Recently, transformers \cite{radosavovic2023realworldhumanoidlocomotionreinforcement} have been used for humanoid locomotion, relying on next-token prediction. In addition, an end-to-end vision-based whole-body control parkour policy \cite{zhuang2024humanoid} has been proposed to overcome multiple parkour skills without any motion prior. IL can address reward design challenges by learning from human demonstrations. Advanced techniques like Generative Adversarial Imitation Learning (GAIL) ~\cite{ho2016generative} allow for more adaptable skills from unstructured data. An IL framework using adversarial motion priors has been proposed to demonstrate human-like locomotion. In addition, \citet{radosavovic2024humanoid} constructed a large locomotion dataset, and a transformer was employed to imitate the demonstrations within it.

\section{Method}

As illustrated in Figure \ref{fig:method_framework}, the framework of \method consists of three main components: a retargeting network, a dual-level controller, and the curriculum learning procedure. Our whole-body controller supports two command modes: 1) \textbf{root-based mode}, which includes root velocity commands and reference joint angle commands for the upper body, and 2) \textbf{pose-based mode}, which provides reference joint angle commands for the entire body. Specifically, the root velocity command includes linear forward and sideways velocities, as well as angular turning velocity.

\subsection{Regtargeting Network}

Retargeting is an essential data preprocessing procedure. At the same time, real-time retargeting is also crucial for low-latency teleoperation. \textbf{In this section, we want to propose a different perspective on retargeting for low-latency motion adaptation.} Specifically, we adopt deep mimic methods instead of directly using traditional IK techniques.

We first we adopt a two-stage optimization approach as H2O \citep{he2024learning} to process the AMASS dataset \citep{AMASS:ICCV:2019}, generating a large set of human and humanoid pose pairs. Details are available in the Appendix. Then, a lightweight three-layer MLP is employed to learn the mapping between them. This approach significantly improves the real-time performance of the retargeting process while maintaining pose transfer accuracy. Unlike optimization methods used in prior works, MLP ensures smooth output even with small noise perturbations on input, eliminating frame-to-frame jitter, which is verified in Section \ref{retargeting_analysis}. This is because the MLP is a continuous nonlinear function approximator and trained on massive filtered motion data. Specifically, we manually remove frames with discontinuous motions resulting from unstable IK solutions. Note that, for the data preprocessing procedure, we retarget the motion sequences from the AMASS dataset and also apply the same filtering process as HumanPlus \cite{fu2024humanplus} to ensure high-quality retargeting data. 

\setcounter{figure}{1}
\begin{figure*}[t]
    \centering
    \includegraphics[width=.95\linewidth]{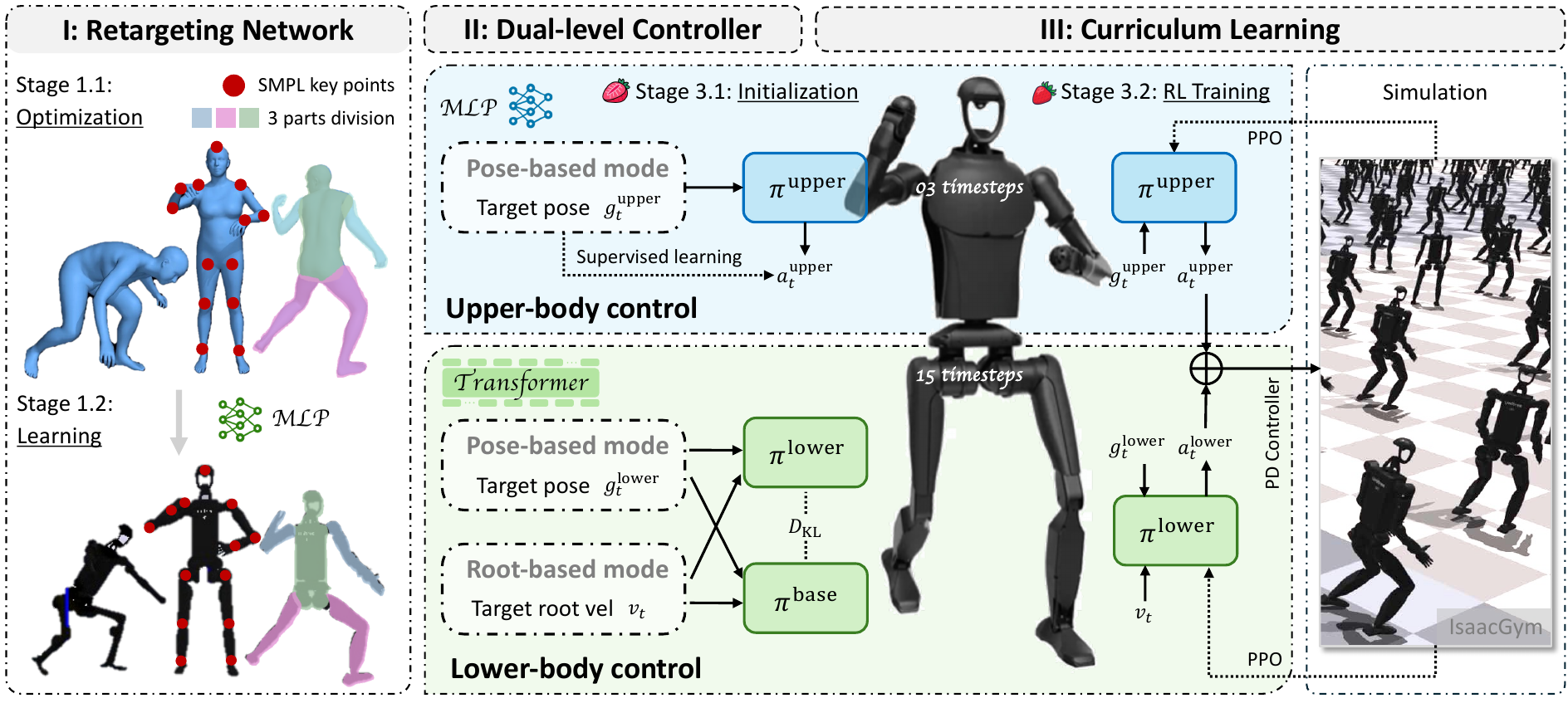}
    \vspace{2mm}
    \caption{
    The framework of \method. The left shows the retargeting network, which uses an MLP to learn the relationship between human and robot pose pairs. The middle section illustrates how we control the upper and lower body parts using two separate controllers. The right section presents the curriculum learning paradigm, where controllers are first trained using supervised learning and then improved through RL to explore a better policy.
}
    \label{fig:method_framework}
    \vspace{-.2cm}
\end{figure*}

\subsection{Dual-level Controller}

\method aims to track the reference pose and root velocity more precisely and robustly. To this end, we implement a dual-level controller that separately manages the upper and lower bodies. Intuitively, this approach decomposes the action space, thus reducing the learning complexity. In addition, The independent controllers offer greater flexibility in selecting their individual network architectures. 

\textbf{Setting Formulation}: We formulate the humanoid whole-body control problem as a Decentralized Partially Observable Markov Decision Process (Dec-POMDP) \cite{oliehoek2016concise} with two agents. Specifically, one agent controls the lower body using the policy $\pi^{\rm lower}$, and the other one controls the upper body using the policy $\pi^{\rm upper}$. The observation at timestep $t$ consists of proprioceptive state $o_t$, previous action $a_{t-1}=[a_{t-1}^{\rm lower}, a_{t-1}^{\rm upper}]$, root velocity tracking target $v_t$, and the pose tracking target $g_t$. Both policies take as input a \(n\)-step observation history \(\boldsymbol{x}_{t,n} = \{ \boldsymbol{o}_{t-n:t}, \boldsymbol{a}_{t-n-1:t-1}, \boldsymbol{v}_{t-n:t}, \boldsymbol{g}_{t-n:t} \} \), and output action $a^{\rm lower}_t$ and $a^{\rm upper}_t$, respectively. The predicted actions are then concatenated to form the target joint positions of joint proportional derivative (PD) controllers.

\textbf{Lower-body Controller}:
The task of the lower body is particularly challenging, as it not only needs to track the root velocity and reference pose, but also must maintain overall body balance simultaneously. Recent works \cite{radosavovic2024humanoid, radosavovic2023realworldhumanoidlocomotionreinforcement} have demonstrated that in locomotion tasks on complex terrains requiring high balance control, Transformer \citep{vaswani2017attention} outperforms MLP and LSTM \citep{schmidhuber1997long}. These results suggest that the history of observations and actions implicitly encodes environmental information, enabling the Transformer model to dynamically adapt its behavior. Inspired by this, we instantiate the lower-body controller as a Transformer, given that WBC is a more complex, long-range task.

Specifically, we adopt the Gated Transformer-XL architecture \cite{parisotto2020stabilizing}, which consists of three layers of transformer blocks. This Transformer variant is designed to better suit RL tasks by replacing standard residual connections with gating mechanisms within the transformer block. These gates can substantially improve the learning stability and speed in RL.

\textbf{Upper-body Controller}: We implement it as a lightweight 3-layer MLP and it is sufficient since it only needs to follow the reference pose without complex balance control. As shown in experiments, MLP achieves the desired functionality with minimal computational cost.

\subsection{Curriculum Learning}
Training a root-based WBC controller from scratch with RL is relatively straightforward. However, when simultaneously training both lower-body root velocity tracking and upper-body pose tracking, we observe significant instability. This is likely due to the distinct objectives of the upper and lower bodies, which interfere with each other during training. It highlights the necessity of curriculum learning.

\textbf{Base policy}: Specifically, as training a lower-body controller independently is much easier, we first train it while keeping the upper-body joints fixed, resulting in a base policy $\pi^{\rm base}$. The base policy can either be a locomotion policy or a joint angle tracking policy, depending on whether we aim to train the root-based mode or pose-based mode controller.

\textbf{Supervised Initialization}: We initially attempted to continue training WBC with RL from the base policy, but encountered two problems. The first one is that the upper-body controller is prone to be overly conservative, failing to accurately track the reference pose. The other one is that the unconvergent upper-body controller disrupted the training of the lower-body controller, causing it to deviate from the base policy and compromising its ability to maintain balance. To address these, we propose a supervised initialization process. Firstly, supervised learning is applied to the upper body to directly mimic the reference pose:
\(
    L_{\rm SL}(\pi^{\rm upper}) = \|\pi^{\rm upper}(\boldsymbol{x}_{t,3}) - \boldsymbol{g}^{\rm upper}_t\|.
\)
Meanwhile, we impose a KL regularization on the lower-body controller to prevent a large deviation from the base policy:
\(
    L_{\rm KL}(\pi^{\rm lower}) = D_{\rm KL}\left[\pi^{\rm lower}(\cdot\mid \boldsymbol{x}_{t,15})\|\pi^{\rm base}(\cdot\mid \boldsymbol{x}_{t,15})\right].
\)
While these two objectives may result in a whole-body controller with suboptimal performance, they provide a solid initialization, facilitating faster learning during subsequent RL training and reducing the risk of converging to overly conservative local optima. In addition, to further simplify the task, we reduce the pose amplitude to half of its original value.

\textbf{RL Training}: Based on the initialized policy introduced in the previous section, we further improve its performance with RL. At this stage, we adjust the pose magnitude back to its original range. We use the PPO \cite{schulman2017proximal} as the base RL algorithm to maximize the expectation of the accumulated future rewards, \(\mathbb{E}_{\pi^{\rm lower}, \pi^{\rm upper}} \left[ \sum_{t=0}^{T} \gamma^t r(s_t, a_t) \right] \), where \(r(s_t, a_t)\) represents the reward received by the whole-body controller after taking action \(a_t\) at state \(s_t\).

\begin{wraptable}{r}{0.55\textwidth}
\vspace{-0.2cm}
\centering
\normalsize
\caption{Rewards Specification for \method.}
\label{tab:rewards_detailed}
\renewcommand{\arraystretch}{1.2}
\small
\begin{tabular}{@{}llr@{}}
\toprule
\textbf{Term} & \textbf{Expression} & \textbf{Weight} \\ \midrule
\multicolumn{3}{c}{\textcolor{upperbody}{\textbf{Upper Body}}} \\ 
\midrule
DoF Position & exp${(-4.0 * |\boldsymbol{q}_{\text{ref}}-\boldsymbol{q}|)}$ & 20.0 \\ \midrule

\multicolumn{3}{c}{\textcolor{lowerbody}{\textbf{Lower Body}}} \\ \midrule
DoF Position & exp$(-4.0 * |\boldsymbol{q}_{\text{ref}}-\boldsymbol{q}|)$ & 100.0 \\
Feet Height &  exp$(-10.0 *|\boldsymbol{h}_{\text{target}}-\boldsymbol{h}|)$ & 100.0 \\
Linear Velocity & exp$(-4.0 * |\boldsymbol{v}_\text{lin\_cmd}-\boldsymbol{v}_\text{lin}|)$ & 20.0 \\
Angular Velocity & exp$(-4.0 * |\boldsymbol{v}_\text{ang\_cmd}- \boldsymbol{v}_\text{ang}|)$ & 20.0 \\
Roll \& Pitch & exp$(-4.0 * |\boldsymbol{\Omega}_\text{ref}^{\phi\theta}-\boldsymbol{\Omega}^{\phi\theta}|)$ & 1.0 \\
\midrule
\end{tabular}
\vspace{-0.2cm}
\end{wraptable}

Our reward function, as listed in Table \ref{tab:rewards_detailed}, is carefully designed to improve both the performance and real-world applicability of the humanoid’s motion. It primarily includes rewards for tracking the root's velocity and orientation, along with precise joint angle tracking. Additionally, we introduce several regularization terms to enhance the robot's stability and improve the transferability from simulation to real-world applications. The key components of our tracking reward are outlined in Table \ref{tab:rewards_detailed}, while supplementary rewards focused on stability and sim-to-real transfer are discussed in detail in the appendix.

After training the models for both modes separately, we distill them into a single network via imitation learning. Note that, for the root-based mode, we only need to set the lower-body reference pose command to zero.

\section{Experiment}

\subsection{Experimental Setup}

We conduct our experiments in IsaacGym \cite{makoviychuk2021isaac} simulator across two adult-size humanoid robot platforms, Unitree 19-DoF H1 and 21-DoF H1-2. We noticed that while G1 performs well on WBC tasks,
the sim-to-real gap is much larger for adult-size robots. The Unitree H1-2 upgraded added DOFs, motors, and computational modules, increasing upper body weight and control difficulty. To the best of our knowledge, \textbf{effective WBC for H1-2 remains an open challenge but is more crucial for real-world applications than G1.}

To examine the effectiveness of our method, we evaluate three \textit{open-sourced SOTA} baselines: \textbf{HumanPlus} \cite{fu2024humanplus}, \textbf{Exbody} \citep{cheng2024express} and \textbf{OmniH2O} \cite{he2024omnih2o}. For the baseline evaluation set, we utilize a subset of the AMASS dataset, denoted as $\mathcal{\hat{Q}}$. Following the curation method described in OmniH2O \cite{cheng2024express}, $\mathcal{\hat{Q}}$ preserves the diversity of movements within the AMASS dataset while ensuring feasibility for humanoid robots. Evaluation errors are measured on $\mathcal{\hat{Q}}$ using five random seeds. Specifically, we adopt mean absolute error (MAE) as the evaluation metric. \textit{It measures the tracking error between the desired motion coming from the retargeting references and the measurements of the robot in the simulation.} To ensure a fair comparison, our retargeting method is applied as a preprocessing step for all baselines.

For the \textbf{root-based mode} evaluation, we use the upper body joint angles from the evaluation set \(\mathcal{\hat{Q}}\) and randomly assign root velocity commands. We evaluate the performance of each method via linear velocity tracking error \(E_\text{root-lin-vel}\) (m/s) and angular velocity tracking error \(E_\text{root-ang-vel}\) (rad/s) for the lower body. For the upper-body pose tracking, we calculate the joint tracking error \( E_\text{upper-j}\) (rad). Note that OmniH2O is excluded from this mode since it does not support root commands.

For the \textbf{pose-based mode} evaluation, we use the whole-body poses from the evaluation set \(\mathcal{\hat{Q}}\) and evaluate the tracking ability of each method. Specifically, we compare the methods based on upper-body and lower-body joint tracking errors, \( E_\text{upper-j} \) and \( E_\text{lower-j} \) (rad), as well as root velocity tracking error \(E_\text{root-vel}\) (m/s) and root orientation tracking error, \(E_\text{root-r}\), \(E_\text{root-p}\) and \(E_\text{root-y}\) (rad). In this mode, the target root velocity and orientation are extracted from each trajectory in the evaluation set. In addition, since OmniH2O is the keypoint position tracking method, we convert its reference keypoint positions to corresponding joint angles to ensure a fair comparison on the same dimensional basis. Note that Exbody is excluded in this mode since it does not support lower-body pose tracking.

\subsection{Main Experiments}

\begin{wraptable}{r}{0.5\textwidth}
\vspace{-4mm}
\caption{Main results of \method in the root-based mode. For root tracking, we compare errors in tracking root linear and angular velocity. For the upper body control, we compare errors in tracking joint angles.}
\label{tab:root_main_result}
\centering
\resizebox{0.5\textwidth}{!}{
\begingroup
\setlength{\tabcolsep}{3pt} 
\renewcommand{\arraystretch}{1.5} 
\hspace{-4mm}
\begin{tabular}{llccc}
\toprule
& & \multicolumn{2}{c}{\textcolor{NavyBlue}{\textbf{Root Tracking Error}}} & \multicolumn{1}{c}{\textcolor{Peach}{\textbf{Joint Angle Error}}} \\ 
\cmidrule(lr){3-4} \cmidrule(lr){5-5}

\multicolumn{2}{l}{\textbf{Method}} & $E_\text{root-lin-vel}\downarrow$ & $E_\text{root-ang-vel}\downarrow$  & $E_\text{upper-j}\downarrow$ \\ 
\midrule

\multirow{3}{*}{\textbf{H1}} & ExBody & $0.2417 \pm \text{\tiny 0.0004}$ & $0.4176 \pm \text{\tiny 0.0002}$ & $0.1689 \pm \text{\tiny 0.0006}$ \\ 

& HumanPlus & $0.1696 \pm \text{\tiny 0.0013}$ & $0.3105 \pm \text{\tiny 0.0021}$ & $0.2187 \pm \text{\tiny 0.0041}$ \\
& \textbf{\methodours} & 
\cellcolor{tablepeach}{\textcolor{black}{$0.1504 \pm \text{\tiny 0.0264}$}}
& \cellcolor{tablepeach}{\textcolor{black}{$0.0937 \pm \text{\tiny 0.0142}$}}
& \cellcolor{tablepeach}{\textcolor{black}{$0.1093 \pm \text{\tiny 0.0010}$}} \\
\midrule 

\multirow{2}{*}{\textbf{H1-2}} 
& HumanPlus  & $0.4026 \pm \text{\tiny 0.0008}$ & $0.6209 \pm \text{\tiny 0.0015}$ & $0.2547 \pm \text{\tiny 0.0009}$ \\
& \textbf{\methodours} & 
\cellcolor{tablepeach}{\textcolor{black}{$0.2224 \pm \text{\tiny 0.0150}$}}& 
\cellcolor{tablepeach}{\textcolor{black}{$0.2184 \pm \text{\tiny 0.0062}$}}& 
\cellcolor{tablepeach}{\textcolor{black}{$0.0933 \pm \text{\tiny 0.0001}$}} \\
\bottomrule
\end{tabular}
\endgroup
}
\vspace{-0.4cm}
\end{wraptable}

\textbf{Root-based mode}: The evaluation results of the root-based mode are presented in Table \ref{tab:root_main_result}.  Specifically, we compare \method with two SOTA baselines, HumanPlus and ExBody, that support root command and release the code. The officially released ExBody checkpoint is used for comparison. For HumanPlus, we reproduced the results using the officially released code. For the root velocity training and evaluation range, we set the linear forward velocity as $[-0.5, 1]$ m/s, linear sideways velocity as $[-0.5, 0.5]$ m/s, and the angular velocity as $[-0.5, 0.5]$ rad/s. Note that we randomly sample root commands from the evaluation range and the upper-body reference pose from the evaluation set.

For both H1 and H1-2, \method significantly outperforms the baseline in both velocity tracking error and upper-body joint angles tracking error. In more detail, for H1, the average error of the upper body is approximately 0.1 radians (5.7 degrees), while the errors of the other methods are at least 10 degrees. For velocity tracking error, \method also shows a significant improvement, especially in angular velocity tracking, with an error of only 5 degrees per second. HumanPlus, due to its single-network structure, exhibits noticeable deficiencies in both upper-body angle tracking and angular velocity tracking. Surprisingly, ExBody, using the official checkpoint, performs the worst. This may be because its training set consists of only a small subset of AMASS. However, both \method and HumanPlus utilize the entire dataset, resulting in stronger generalization capabilities. For H1-2, we can draw similar experimental conclusions as H1. However, in velocity tracking, H1-2 performs worse than H1 due to its heavier upper body (with heavier arms and computing modules), which increases the difficulty of lower-body training.

\textbf{Pose-based mode}: The evaluation results of the pose-based mode are presented in Table \ref{tab:exp_table}. Specifically, we compare \method with two SOTA baselines, HumanPlus and OmniH2O, which support tracking reference pose. We reproduced their results using the officially released code. For both H1 and H1-2, \method achieves higher tracking accuracy in both upper-body and lower-body pose tracking compared to baselines, HumanPlus and OmniH2O. In detail, the joint angle tracking errors of \method are only 30-60\% of those observed in other methods. Meanwhile, \method also demonstrates comparable accuracy in root tracking. For example, \method exhibits the lowest tracking errors for root roll, pitch, and yaw on H1. This suggests that our method does not sacrifice other tracking capabilities to achieve precise joint angle tracking.

\begin{table*}[tbp]
\caption{Main results of \method in the pose-based mode. 
\method demonstrates significantly lower joint tracking errors than baselines.}
\label{tab:exp_table}
\centering
\resizebox{.95\textwidth}{!}{
\begingroup
\setlength{\tabcolsep}{3pt} 
\renewcommand{\arraystretch}{1.5} 
\begin{tabular}{llccccccc}
\toprule
\multicolumn{2}{c}{} & \multicolumn{5}{c}{\textcolor{NavyBlue}{\textbf{Root Tracking Error}}} & \multicolumn{2}{c}{\textcolor{Peach}{\textbf{Joint Angle Error}}} \\ 
\cmidrule(lr){3-7} \cmidrule(lr){8-9}

\multicolumn{2}{l}{\textbf{Method}} & $E_\text{root-vel}\downarrow$ & $E_\text{root-height}\downarrow$ & $E_\text{root-r}\downarrow$ & $E_\text{root-p}\downarrow$ &$E_\text{root-y}\downarrow$ & $E_\text{upper-j}\downarrow$ & $E_\text{lower-j}\downarrow$\\ 
\midrule

\multirow{3}{*}{\textbf{H1}} & HumanPlus& 
$0.4220 \pm \text{\tiny 0.0047}$ & 
{\textcolor{black}{$0.1015 \pm \tiny{\text{0.0004}}$}} & 
$0.0766 \pm \text{\tiny 0.0006}$ & $0.1137 \pm \text{\tiny 0.0003}$ & 
$0.6147 \pm \text{\tiny 0.0092}$ & $0.2199 \pm \text{\tiny 0.0047}$ & 
$0.1284 \pm \text{\tiny 0.0002}$ \\ 

& OmniH2o & 
$0.5424 \pm \text{\tiny 0.0028}$ & 
\cellcolor{tablepeach}{\textcolor{black}{$0.0853 \pm \tiny{\text{0.0022}}$}} & 
$0.0481 \pm \text{\tiny 0.0024}$ & $0.1301 \pm \text{\tiny 0.0041}$ & 
$0.6547 \pm \text{\tiny 0.0067}$ & $0.2217 \pm \text{\tiny 0.0024}$ & 
$0.2404 \pm \text{\tiny 0.0021}$ \\ 

& \textbf{\methodours} & 
\cellcolor{tablepeach}{\textcolor{black}{$0.3181 \pm \text{\tiny 0.0009}$}} & 
$0.1002 \pm \text{\tiny 0.0001}$ & 
\cellcolor{tablepeach}{\textcolor{black}{$0.0415 \pm \text{\tiny 0.0001}$}} & 
\cellcolor{tablepeach}{\textcolor{black}{$0.0595 \pm \tiny{\text{0.0001}}$}} & 
\cellcolor{tablepeach}{\textcolor{black}{$0.5452 \pm \text{\tiny 0.0028}$}} & 
\cellcolor{tablepeach}{\textcolor{black}{$0.0686 \pm \tiny{\text{0.0001}}$}} & 
\cellcolor{tablepeach}{\textcolor{black}{$0.0658 \pm \tiny{\text{0.0001}}$}} \\

\cmidrule(lr){1-2} \cmidrule(lr){3-7} \cmidrule(lr){8-9} 
\multirow{3}{*}{\textbf{H1-2}} & HumanPlus&
\cellcolor{tablepeach}{\textcolor{black}{$0.3078 \pm\text{\tiny 0.0056} $}} &
$0.0673 \pm\text{\tiny 0.0003} $&
$0.0672 \pm\text{\tiny 0.0047}$ &
$0.1270 \pm\text{\tiny 0.0012}$ &
$0.8293 \pm\text{\tiny 0.0064}$ &
$0.2277 \pm\text{\tiny 0.0004}$ &
$0.1722 \pm\text{\tiny 0.0627}$ \\ 

& OmniH2o& $0.4509 \pm \text{\tiny 0.0029}$ & 
{\textcolor{black}{$0.0688 \pm \tiny{\text{0.0021}}$}} & 
\cellcolor{tablepeach}{\textcolor{black}{$0.0314 \pm \text{\tiny 0.0024}$}} & 
\cellcolor{tablepeach}{\textcolor{black}{$0.0515 \pm \text{\tiny 0.0056}$}}& 
\cellcolor{tablepeach}{\textcolor{black}{$0.6540 \pm \text{\tiny 0.0141}$}} & 
$0.2552 \pm \text{\tiny 0.0032}$ & 
$0.2790 \pm \text{\tiny 0.0078}$ \\ 

& \textbf{\methodours} & 
$0.3418 \pm \tiny{\text{0.0004}}$ & 
\cellcolor{tablepeach}{\textcolor{black}{$0.0608 \pm \text{\tiny 0.0000}$}}& 
$0.0398 \pm \text{\tiny 0.0001}$ & 
$0.0931 \pm \text{\tiny 0.0001}$ & 
$0.8051 \pm \tiny{\text{0.0089}}$ & 
\cellcolor{tablepeach}{\textcolor{black}{$0.1192 \pm \tiny{\text{0.0002}}$}} & 
\cellcolor{tablepeach}{\textcolor{black}{$0.1000 \pm \tiny{\text{0.0001}}$}} \\

\bottomrule
\end{tabular}
\endgroup
}
\vspace{-0.4cm}
\end{table*}

Consistent with the results in the root-based mode, we also observe that H1-2 exhibits larger tracking errors due to its heavier body across all methods. While both HumanPlus and \method are designed to track joint angles, \method achieves lower tracking errors across most evaluation metrics, suggesting the effectiveness of our proposed dual-level controller and curriculum learning process. Furthermore, \method exhibits low tracking errors in tracking root rotation and velocity, demonstrating its ability to maintain balance effectively while achieving superior lower-body pose tracking performance compared to the baselines. OmniH2O demonstrates certain advantages in root tracking, as it is trained to follow global keypoint positions, whereas HumanPlus and \method are designed to track joint angles. As a result, OmniH2O shows the worst tracking ability in joint angles.

\textbf{Component Ablations}: We conduct ablation studies to evaluate the effectiveness of each component. Specifically, we compare \method with two ablation baselines: one without supervised initialization, referred to as \textbf{\method w.o. Init}, and another replacing the dual-level controller with a single transformer network to control the whole body, referred to as \textbf{\method single}.

We conduct ablation studies to evaluate the effectiveness of each component. Specifically, we compare \method with two ablation baselines: one without supervised initialization, referred to as \textbf{\method w.o. Init}, and another replacing the dual-level controller with a single transformer network to control the whole body, referred to as \textbf{\method single}.

\begin{wraptable}{r}{0.5\textwidth}
\caption{Ablation studies of \method in root-based mode. \method w.o. Init refers to the controller trained directly without supervised initialization. \method Single refers to the controller using one transformer network to control the whole body.}
\label{tab:root_ablation}
\centering
\resizebox{0.5\textwidth}{!}{
\begingroup
\setlength{\tabcolsep}{3pt} 
\renewcommand{\arraystretch}{1.5} 
\hspace{-4mm}
\begin{tabular}{llccc}
\toprule
& & \multicolumn{2}{c}{\textcolor{NavyBlue}{\textbf{Root Tracking Error}}} & \multicolumn{1}{c}{\textcolor{Peach}{\textbf{Joint Angle Error}}} \\ 
\cmidrule(lr){3-4} \cmidrule(lr){5-5}

\multicolumn{2}{l}{\textbf{Method}} & $E_\text{root-lin-vel}\downarrow$ & $E_\text{root-ang-vel}\downarrow$  & $E_\text{upper-j}\downarrow$ \\ 
\midrule

\multirow{3}{*}{\textbf{H1}} & \method w.o. Init & $0.1953 \pm \text{\tiny 0.0152}$ & $0.2015 \pm \text{\tiny 0.0075}$ & $0.1627 \pm \text{\tiny 0.0001}$\\ 
& \method Single & $0.2039 \pm \text{\tiny 0.0010}$ & $0.2776 \pm \text{\tiny 0.0402}$ & $0.5377 \pm \text{\tiny 0.0003}$\\
& \textbf{\methodours} & 
\cellcolor{tablepeach}{\textcolor{black}{$0.1504 \pm \text{\tiny 0.0264}$}}
& \cellcolor{tablepeach}{\textcolor{black}{$0.0937 \pm \text{\tiny 0.0142}$}}
& \cellcolor{tablepeach}{\textcolor{black}{$0.1093 \pm \text{\tiny 0.0010}$}} \\
\midrule 

\multirow{3}{*}{\textbf{H1-2}} & \method w.o. Init & $0.2488 \pm \text{\tiny 0.0236}$ & $0.2780 \pm \text{\tiny 0.0185}$ & $0.1286 \pm \text{\tiny 0.0091}$ \\ 
& \method Single & 
\cellcolor{tablepeach}{\textcolor{black}{$0.2026 \pm \text{\tiny 0.0273}$}}
& \cellcolor{tablepeach}{\textcolor{black}{$0.1865 \pm \text{\tiny 0.0230}$}}
& $0.4693 \pm \text{\tiny 0.0070}$\\
& \textbf{\methodours} & $0.2224 \pm \text{\tiny 0.0150}$ & $0.2184 \pm \text{\tiny 0.0062}$ & 
\cellcolor{tablepeach}{\textcolor{black}{$0.0933 \pm \text{\tiny 0.0001}$}}\\

\bottomrule
\end{tabular}
\endgroup
}
\vspace{-2mm}
\end{wraptable}

In the root-based mode, we found that while \method slightly outperforms the two ablation methods in velocity tracking overall, there is a significant difference in upper-body tracking. As shown in Table \ref{tab:root_ablation}, \method performs the best in upper-body tracking, whereas \method Single performs the worst. We speculate that when a single policy controls the entire body, upper-body and lower-body movements may interfere with each other. Specifically, larger upper-body movements can make it more difficult to maintain lower-body stability. During training, the policy may sacrifice the performance of one task to improve the tracking ability of the other. For example, on H1-2, \method Single achieves the best performance in root tracking error but the worst in joint angle tracking. This highlights the advantage of the dual-level controller, where separate controllers can focus more effectively on their respective tasks. Additionally, \method's superior performance compared to \method w.o. Init emphasizes the importance of supervised initialization, as good initialization influences the model’s ability to converge to an optimal solution.

As shown in Table \ref{tab:pose_ablation}, a similar trend is observed in the pose-based mode: the dual-level controller outperforms the single controller, especially in upper-body joint angle tracking, and supervised initialization improves model convergence and performance. Similar to the root-based mode, \method Single struggles with upper-body joint tracking but performs well in lower-body joint tracking, maintaining low errors. Additionally, \method w.o. Init exhibits significant failure in upper-body tracking for H1-2. This is due to poor initialization, which may cause some joints in the upper body to converge to a suboptimal solution, resulting in a higher average error. As for lower-body tracking, the three methods demonstrate comparable performance. Differences in root tracking errors among the three methods are less significant compared to the root-based mode. This can be attributed to the generally lower root velocity in the pose-based mode, which reduces the interference of upper-body movements with lower-body balance. In conclusion, these findings further confirm the effectiveness of each component.

\begin{table*}[tbp]
\caption{Ablation studies of \method in pose-based mode. \method w.o. Init refers to the controller trained directly without supervised initialization. \method Single refers to the controller using one transformer network to control the whole body.}
\label{tab:pose_ablation}
\centering
\resizebox{.95\textwidth}{!}{
\begingroup
\setlength{\tabcolsep}{3pt} 
\renewcommand{\arraystretch}{1.5} 
\begin{tabular}{llccccccc}
\toprule
\multicolumn{2}{c}{} & \multicolumn{5}{c}{\textcolor{NavyBlue}{\textbf{Root Tracking Error}}} & \multicolumn{2}{c}{\textcolor{Peach}{\textbf{Joint Angle Error}}} \\ 
\cmidrule(lr){3-7} \cmidrule(lr){8-9}

\multicolumn{2}{l}{\textbf{Method}} & $E_\text{root-vel}\downarrow$ & $E_\text{root-height}\downarrow$ & $E_\text{root-r}\downarrow$ & $E_\text{root-p}\downarrow$ &$E_\text{root-y}\downarrow$ & $E_\text{upper-j}\downarrow$ & $E_\text{lower-j}\downarrow$\\ 
\midrule

\multirow{3}{*}{\textbf{H1}} & \method w.o. Init & 
\cellcolor{tablepeach}{\textcolor{black}{$0.3059 \pm \text{\tiny 0.0009}$}} & 
$0.0982 \pm \text{\tiny 0.0000}$ & 
$0.0419 \pm \text{\tiny 0.0001}$ & 
\cellcolor{tablepeach}{\textcolor{black}{$0.0549 \pm \text{\tiny 0.0001}$}} & 
$0.5718 \pm \text{\tiny 0.0040}$ &
$0.0780 \pm \text{\tiny 0.0001}$ & 
$0.0667 \pm \text{\tiny 0.0001}$ \\ 

& \method Single & 
$0.3142 \pm \text{\tiny 0.0009}$ & 
\cellcolor{tablepeach}{\textcolor{black}{$0.0958 \pm \text{\tiny 0.0001}$}} & 
\cellcolor{tablepeach}{\textcolor{black}{$0.0408 \pm \tiny{\text{0.0001}}$}} &
$0.0550 \pm \text{\tiny 0.0001}$ & 
$0.5601 \pm \text{\tiny 0.0010}$ & 
$0.2469 \pm \text{\tiny 0.0001}$ & 
\cellcolor{tablepeach}{\textcolor{black}{$0.0654 \pm \text{\tiny 0.0001}$}} \\

& \textbf{\methodours} & 
\textcolor{black}{$0.3181 \pm \text{\tiny 0.0009}$} & 
$0.1002 \pm \text{\tiny 0.0001}$ & 
\textcolor{black}{$0.0415 \pm \text{\tiny 0.0001}$} & 
$0.0595 \pm \tiny{\text{0.0001}}$ & 
\cellcolor{tablepeach}{\textcolor{black}{$0.5452 \pm \text{\tiny 0.0028}$}} & 
\cellcolor{tablepeach}{\textcolor{black}{$0.0686 \pm \tiny{\text{0.0001}}$}} & 
$0.0658 \pm \tiny{\text{0.0001}}$ \\

\cmidrule(lr){1-2} \cmidrule(lr){3-7} \cmidrule(lr){8-9} 
\multirow{3}{*}{\textbf{H1-2}} & \method w.o. Init & 
$0.4108 \pm \text{\tiny 0.0037}$ & 
$0.0664 \pm \text{\tiny 0.0045}$ & 
$0.0577 \pm \text{\tiny 0.0005}$ & 
\cellcolor{tablepeach}{\textcolor{black}{$0.0781 \pm \text{\tiny 0.0002}$}} & 
$1.5397 \pm \text{\tiny 0.0089}$ & 
$0.4088 \pm \tiny{\text{0.0010}}$ & 
$0.1137 \pm \tiny{\text{0.0004}}$ \\

& \method Single &
$0.6634 \pm \text{\tiny 0.0028}$ & 
$0.0643 \pm \text{\tiny 0.0003}$ & 
$0.0755 \pm \text{\tiny 0.0002}$ & 
$0.0989 \pm \text{\tiny 0.0002}$ & 
$1.1690 \pm \text{\tiny 0.0099}$ & 
$0.2477 \pm \tiny{\text{0.0005}}$ & 
$0.1112 \pm \tiny{\text{0.0003}}$ \\

& \textbf{\methodours} & 
\cellcolor{tablepeach}{\textcolor{black}{$0.3418 \pm \tiny{\text{0.0004}}$}} & 
\cellcolor{tablepeach}{\textcolor{black}{$0.0608 \pm \text{\tiny 0.0000}$}} & 
\cellcolor{tablepeach}{\textcolor{black}{$0.0398 \pm \text{\tiny 0.0001}$}} & 
$0.0931 \pm \text{\tiny 0.0001}$ & 
\cellcolor{tablepeach}{\textcolor{black}{$0.8051 \pm \tiny{\text{0.0089}}$}} & 
\cellcolor{tablepeach}{\textcolor{black}{$0.1192 \pm \tiny{\text{0.0002}}$}} & 
\cellcolor{tablepeach}{\textcolor{black}{$0.1000 \pm \tiny{\text{0.0001}}$}} \\

\bottomrule
\end{tabular}
\endgroup
}
\vspace{-0.4cm}
\end{table*}

\subsection{Retargeting Analysis}
\label{retargeting_analysis}
In this part, we evaluate our retargeting method and baselines from the perspectives of real-time performance, accuracy, and smoothness. Since retargeting does not need to consider whether the pose is feasible for the humanoid, we use the entire AMASS dataset as the evaluation set. We conduct the following experiments on a platform equipped with an RTX 4060 GPU. 

\begin{figure}[t]
    \centering
    \begin{minipage}{0.35\textwidth}
        \captionof{table}{The comparison of different retargeting methods in terms of real-time performance ($\rm s/frame$) and retargeting error ($\rm m/frame$).}
        \renewcommand{\arraystretch}{1.5}
        \resizebox{\linewidth}{!}{
            \begin{tabular}{lcc}
            \toprule
            \textbf{Method} & \textbf{Real-time} & \textbf{Error} \\
            \midrule
            HumanPlus & 0.00200 & 0.16908 \\
            H2O & 9.05409 & 0.14476\\
            \methodours & 0.00064 & 0.13567 \\
            \bottomrule
            \end{tabular}
        }
        \label{table_retargeting}
    \end{minipage}
    \hfil
    \begin{minipage}{0.59\textwidth}
        \captionof{table}{The smoothness of the joint angle sequence is evaluated using the Root Mean Square of angular acceleration, the second difference of the angle ($\rm rad/s^2$), where lower values indicate higher smoothness.}
        \resizebox{\linewidth}{!}{
            \renewcommand{\arraystretch}{1.5}
            \begin{tabular}{lcccccc}
                \toprule
                & \multicolumn{3}{c}{\textbf{Left Shoulder}} & \multicolumn{3}{c}{\textbf{Right Shoulder}} \\
                \cmidrule(lr){2-4} \cmidrule(lr){5-7}
                \textbf{Method} & \textbf{Pitch} & \textbf{Roll} & \textbf{Yaw} & \textbf{Pitch} & \textbf{Roll} & \textbf{Yaw} \\
                \midrule
                H2O  & 0.0143 & 0.0105 & 0.0147 & 0.0100 & 0.0065 & 0.0118 \\
                \methodours & 0.0069 & 0.0033 & 0.0052 & 0.0070 & 0.0031 & 0.0054 \\
                \bottomrule
            \end{tabular}
        }
        \label{tab:smoothness_wrap}
    \end{minipage}
\end{figure}

For real-time performance, each retargeting method processes motion sequences frame by frame, with the metric being the average time taken to process a single frame. The results are shown in Table \ref{table_retargeting}. Our method significantly outperforms the others. It turns out that the MLP enables control rates over 1000Hz, which is ideal for low-latency teleoperation applications, \eg surgical robotics. For accuracy, the targets are joint positions of the human original pose in terms of its shape. We compare the average error between the joint positions of the humanoid robot obtained by different methods and the targets. As shown in Table \ref{table_retargeting}, our method achieves accuracy comparable to H2O and outperforms HumanPlus. The subpar performance of HumanPlus is due to directly copying the angles without considering the differences in DoF structures and geometric sizes between the humanoid and the human model.

We also want to emphasize that MLP-based retargeting method offers advantages over IK approaches in terms of global continuity and computational efficiency. We randomly select a sequence consisting of 2000 frames. This sequence is then separately processed through our MLP-based method and the IK method adopted in H2O for the retargeting task. Table \ref{tab:smoothness_wrap} shows the smoothness of joint angles of both the left and right shoulders. Specifically, the smoothness is evaluated by the Root Mean Square of angular acceleration. The results verify the superior output smoothness achieved by our MLP-based method, which is essential for real-world deployment.

\section{Conclusion}

We propose a dual-level humanoid whole-body controller that decouples upper and lower body control to enhance training efficiency and reduce interference. A real-time retargeting network improves joint angle accuracy and smoothness, while curriculum learning with supervised initialization ensures effective reinforcement learning. Our method outperforms baselines in simulation and shows promising real-world performance.

\section{Limitations}
\method does not support a wider range of commands, such as keypoint positions. Additionally, some complex poses, such as those requiring one foot to stay off the ground for an extended period or poses with large amplitudes, cannot be deployed on a real robot, even though they may be feasible in simulation. Lastly, the current framework requires extensive reward engineering to prevent undesired behaviors and address the sim-to-real challenge.



\bibliography{example}  

\clearpage

\appendix

\section{Environment Details}

\subsection{Real-world Deployment}

Our real robot is built on the Unitree H1-2 platform, with two onboard Jetson AGX serving as the primary computing and communication unit. The control policy outputs the desired joint positions for each motor and transmits commands to the robot’s low-level interface. The policy runs at an inference frequency of 50 Hz. The low-level interface operates at 500 Hz, ensuring smooth real-time control. Communication between the control policy and the low-level interface is facilitated via Lightweight Communications and Marshalling (LCM) \cite{5649358}.

\subsection{Input and Output Information}

In this section, we will provide detailed input and output information of the controller. Note that the Unitree H1 is a full-size general-purpose humanoid robot developed by Unitree Robotics, featuring multiple high-performance joints. Its legs have 10 DoF, the torso has 1 DoF, and each arm has 8 DoF. The legs H1-2 have 12 DoF and the other DoF is the same as H1.

\textbf{Controller Input}: As shown in Table \ref{tab:controller_input}, the controller's observation of each timestep comprises several components: the scaled base angular velocity (3 dimensions), the projected gravity vector (3 dimensions), the tracking command information (5 dimensions), the scaled difference between the current and default joint positions (21 dimensions for H1-2 and 19 dimensions for H1), the scaled joint velocity (21 dimensions for H1-2 and 19 dimensions for H1), and the previous actions executed by the controller (21 dimensions for H1-2 and 19 dimensions for H1). Lastly, we have reference pose (joint angles) information (21 dimensions for H1-2 and 19 dimensions for H1).

For the lower-body controller, we take 15 timesteps of observation as input. For the upper-body controller, we take 3 timesteps of observation as input.  

\begin{table}[h]
    \centering
    \caption{Controller Observation Components and Their Dimensions}
    \vspace{2mm}
    \resizebox{0.65\textwidth}{!}{
    \renewcommand{\arraystretch}{1.3} 
    \begin{tabular}{lcc}
        \toprule
        \textbf{Input Category} & \textbf{H1-2 Dimensions} & \textbf{H1 Dimensions} \\
        \midrule
        Base Angular Velocity  & 3  & 3  \\
        Projected Gravity Vector  & 3  & 3  \\
        Tracking Command   & 5  & 5  \\
        Joint Positions  & 21  & 19  \\
        Joint Velocity  & 21  & 19  \\
        Previous Actions   & 21  & 19  \\
        Reference Joint Angles   & 21  & 19  \\
        \bottomrule
    \end{tabular}
    }
    \label{tab:controller_input}
\end{table}

\textbf{Critic Input}: The critic's input includes one additional component, the base linear velocity (3 dimensions), compared to the controller's input. All other components remain the same as the controller's input.

\textbf{Tracking Command Information}: For the root-based mode, there are two kinds of input, root velocity command and upper-body reference joint angles. The former one is 3 dimensions, including the linear forward velocity, linear sideways velocity, and the angular velocity. For the training range, we set the linear forward velocity as $[-0.5, 1]$ m/s, linear sideways velocity as $[-0.5, 0.5]$ m/s, and the angular velocity as $[-0.5, 0.5]$ rad/s. 

The latter one is 9-dimension reference joint angle information for both H1-2 and H1. The reference joint angle for the lower body is set to zero. Since the requirements for locomotion capabilities are high, we additionally introduce two dimensions to represent the clock signal, which is encoded using sine and cosine functions following the previous work \cite{gu2024humanoid}.

For the pose mode, the main target information is the reference joint angles. The reference joint angles are for the whole body of the humanoid. For H1 and H1-2, it is 19 dimensions and 21 dimensions respectively. In addition, the root velocity command is also given but it is extracted from the reference motion trajectory. Since the clock signal is meaningless for pose mode, we change it to the target root pitch and roll information which are also from the reference motion sequences.

\textbf{Action Space}: The action is the target position of joint proportional derivative (PD) controllers, which is 21 dimensions for Unitree H1-2 and 19 dimensions for Unitree H1. 

\section{Model Details}

\subsection{Upper-body Controller}

The upper-body controller is a 3-layer MLP, which is the same as the Unitree codebase. More details can be found in \href{https://github.com/unitreerobotics/unitree_rl_gym}{https://github.com/unitreerobotics}. The main difference is that the unitree codebase only take the observation of current timestep as input, while we concatenate 3-timestep observation as input. 

\subsection{Lower-body Controller}

The lower-body controller adopts the Gated Transformer-XL architecture. For code details, please refer to \href{https://github.com/datvodinh/ppo-transformer}{https://github.com/datvodinh/ppo-transformer}

The Gated Transformer XL is designed for processing sequential data by integrating Transformer mechanisms with gated GRUs to enhance information flow. Below is an overview of its main components:
1. SinusoidalPE (Relative Positional Encoding) computes sinusoidal-based relative positional embeddings and uses fixed frequency transformations to provide relative positional awareness to the model.
2. TransformerBlock (Core Unit) consists of Multi-Head Attention, Layer Normalization, GRU-based Gating Mechanisms (GRUGate), and a Feedforward Network. The two GRUGates allow dynamic information flow between layers, improving memory retention.
3. GatedTransformerXL (Overall Architecture) incorporates SinusoidalPE for position encoding to maintain temporal dependencies and employs multiple TransformerBlocks for feature extraction while maintaining long-term memory. For our controller, the configuration of model can be found in Table \ref{tab:transformer_params}. In addition, we take 15-timestep observation as input. 

This architecture is well-suited for tasks requiring long-range dependency modeling, such as trajectory modeling in reinforcement learning, language modeling, or sequential decision-making tasks.

\begin{table}[h]
    \centering
    \caption{Transformer Architecture Parameters}
    \vspace{2mm}
    \resizebox{0.45\textwidth}{!}{ 
    \renewcommand{\arraystretch}{1.3} 
    \begin{tabular}{cc}
    
        \toprule
        \textbf{Parameter} & \textbf{Value} \\
        \midrule
        Number of Blocks (\texttt{num\_blocks}) & 2 \\
        Number of Heads (\texttt{num\_heads}) & 6 \\
        GRU Bias (\texttt{gru\_bias}) & 0.0 \\
        Hidden Size (\texttt{hidden\_size}) & 128 \\
        Embedding Dimension (\texttt{embed\_dim}) & 384 \\
        Memory Length (\texttt{memory\_length}) & 15 \\
        \bottomrule
    \end{tabular}}
     \label{tab:transformer_params}
\end{table}

\section{Training Details}

\textbf{Supervised Initalization}: For supervised learning, we use the standard MSE loss. It measures the mean squared error (squared L2 norm) between the action and target. The target is the reference joint angles. 

For KL regularization, we minimize the Kullback-Leibler divergence (KL divergence) between two policies \( \pi^{\rm lower} \) and \( \pi^{\rm base} \) is given by:
\begin{align}
     &D_{\text{KL}}(\pi^{\rm lower} \parallel \pi^{\rm base}) \nonumber \\
    =& \mathbb{E}_{s \sim \rho_{\pi^{\rm lower}}} \left[ \sum_{a} \pi^{\rm lower}(a|s) \log \frac{\pi^{\rm lower}(a|s)}{\pi^{\rm base}(a|s)} \right]
\end{align}

\textbf{RL training}: The controller is further trained with standard PPO \cite{schulman2017proximal} with supervised initialization. We provide the detailed training hyperparameters in Table \ref{tab:hyperparameters}.

\begin{table}[h]
    \centering
    \caption{Hyperparameters for RL Controller}
    \vspace{2mm}
    \resizebox{0.35\textwidth}{!}{
    \renewcommand{\arraystretch}{1.2}
    
    \begin{tabular}{cc}
        \toprule
        \textbf{Hyperparameter} & \textbf{Value} \\
        \midrule
        Optimizer & Adam \\
        
        \( \beta_1, \beta_2 \) & 0.9, 0.999 \\

        Learning Rate & \( 1 \times 10^{-4} \) \\
        Batch Size & 4096 \\
        Discount factor (\( \gamma \)) & 0.99 \\
        Clip Param & 0.2 \\
        Entropy Coef & 0.01 \\
        Max Gradient Norm & 1 \\
        Learning Epochs & 2 \\
        Mini Batches & 4 \\
        Value Loss Coef & 1 \\

        \bottomrule
    \end{tabular}}
    
    \label{tab:hyperparameters}
\end{table}

\textbf{Reward Design}: In the main text, we provided an overview of our tracking-based reward design. Additionally, our reward function includes various penalties and regularization terms. The components of the regularization reward, along with their respective weights for calculating the final rewards, are detailed in Table \ref{tab:regularization_rewards}. The final reward integrates both the regularization terms and the tracking-based reward to effectively train a robust RL policy.

\begin{table}[h]
    \centering
    \caption{Reward components and weights: penalty rewards for preventing undesired behaviors for sim-to-real transfer and regularization to refine motion.}
    \caption{Hyperparameters for RL Controller}
    \vspace{2mm}
    \resizebox{0.55\textwidth}{!}{
    \renewcommand{\arraystretch}{1.2}
    \begin{tabular}{lcr}
        \toprule
        \textbf{Term} & \multicolumn{1}{l}{\textbf{Expression}} & \textbf{Weight} \\
        \midrule
        \multicolumn{3}{c}{\textbf{Penalty}} \\
        \midrule
        Torque limits & $\mathbf{1}(\tau_t \notin [\tau_{\min}, \tau_{\max}])$ & -10 \\
        DoF position limits & $\mathbf{1}(d_t \notin [q_{\min}, q_{\max}])$ & -10 \\
        DoF velocity limits & $\mathbf{1}(\dot{d}_t \notin [\dot{q}_{\min}, \dot{q}_{\max}])$ & -10 \\
        \midrule
        \multicolumn{3}{c}{\textbf{Regularization}} \\
        \midrule
        DoF acceleration & $\|\ddot{d}_t\|_2^2$ & -3e-8 \\

        Lower-body action rate & $\|a_t^{\text{lower}} - a_{t-1}^{\text{lower}}\|_2^2$ & -20 \\
        Upper-body action rate & $\|a_t^{\text{upper}} - a_{t-1}^{\text{upper}}\|_2^2$ & -5 \\
        action smoothness & $\|\dot{a}_t - \dot{a}_{t-1}\|_2^2$ & -10 \\
        Torque & $\|\tau_t\|$ & -0.0001 \\
        Feet air time & $T_{\text{air}} - 0.25$~\cite{rudin2021learning} & 10000 \\
        Feet contact force & $\|F_{\text{feet}}\|_2^2$ & -10 \\
        Stumble & $\mathbf{1}(F_{\text{feet}}^{xy} > 5 \times F_{\text{feet}}^{z})$ & -0.00125 \\
        Orientation & $\|F_{\text{feet}}^{\text{left}}, F_{\text{feet}}^{\text{right}}, F_{\text{root}}^{z}\|$ & -200 \\
        \bottomrule
    \end{tabular}}
    \label{tab:regularization_rewards}
\end{table}

\textbf{Domain Randomizations}: Detailed domain randomization setups are summarized in Table \ref{tab:dm} following the similar setups in OmniH2O \cite{he2024omnih2o}. 

\begin{table}[h]
    \centering
    \caption{The range of randomization for simulated dynamics, external perturbation, and terrain, which are important for sim-to-real transfer, robustness, and generalizability.}
    \vspace{2mm}
    \resizebox{0.4\textwidth}{!}{
    \renewcommand{\arraystretch}{1.2}
    \begin{tabular}{l c}
        \toprule
        \textbf{Term} & \textbf{Value} \\
        \midrule
        \multicolumn{2}{c}{\textbf{Dynamics Randomization}} \\
        \midrule
        Friction & $\mathcal{U}(0.2, 1.1)$ \\
        Base CoM offset & $\mathcal{U}(-0.1, 0)$ m \\
        Link mass & $\mathcal{U}(0.7, 1.3) \times$ default kg \\
        P Gain & $\mathcal{U}(0.75, 1.25) \times$ default \\
        D Gain & $\mathcal{U}(0.75, 1.25) \times$ default \\
        Control delay & $\mathcal{U}(20, 60)$ ms \\
        \midrule
        \multicolumn{2}{c}{\textbf{External Perturbation}} \\
        \midrule
        Push robot & interval = 5s, $v_{xy} = 1$m/s \\
        \midrule
        \multicolumn{2}{c}{\textbf{Randomized Terrain}} \\
        \midrule
        Terrain type & flat, rough \\
        \bottomrule
    \end{tabular}}
    \label{tab:dm}
\end{table}

\end{document}